\documentclass[10pt,twocolumn,letterpaper]{article}

\usepackage{iccv}
\usepackage{times}
\usepackage{epsfig}
\usepackage{graphicx}
\usepackage{amsmath}
\usepackage{amssymb}
\usepackage{makecell}
\usepackage{booktabs}
\usepackage{pifont}

\usepackage[pagebackref=true,breaklinks=true,letterpaper=true,colorlinks,bookmarks=false]{hyperref}

\iccvfinalcopy % *** Uncomment this line for the final submission

\def\iccvPaperID{1673} % *** Enter the ICCV Paper ID here

\ificcvfinal\fi

\begin{document}

%%%%%%%%% TITLE
\title{Dilated SpineNet for Semantic Segmentation}

\author{
Abdullah Rashwan\thanks{Authors contributed equally.}\qquad
Xianzhi Du\footnotemark[1]\qquad
Xiaoqi Yin\qquad
Jing Li\\
Google\\
{\tt\small \{arashwan,xianzhi,yinx,jingli\}@google.com}
}

\maketitle

\ificcvfinal\thispagestyle{empty}\fi

%%%%%%%%% ABSTRACT
\begin{abstract}
Scale-permuted networks have shown promising results on object bounding box detection and instance segmentation. Scale permutation and cross-scale fusion  of features enable the network to capture multi-scale semantics while preserving spatial resolution. In this work, we evaluate this meta-architecture design on semantic segmentation -- another vision task that benefits from high spatial resolution and multi-scale feature fusion at different network stages. By further leveraging dilated convolution operations, we propose SpineNet-Seg, a network discovered by NAS that is searched from the DeepLabv3 system. SpineNet-Seg is designed with a better scale-permuted network topology with customized dilation ratios per block on a semantic segmentation task. SpineNet-Seg models outperform the DeepLabv3/v3+ baselines at all model scales on multiple popular benchmarks in speed and accuracy. In particular, our SpineNet-S143+ model achieves the new state-of-the-art on the popular Cityscapes benchmark at 83.04\% mIoU and attained strong performance on the PASCAL VOC2012 benchmark at 85.56\% mIoU. SpineNet-Seg models also show promising results on a challenging Street View segmentation dataset. Code and checkpoints will be open-sourced.

\end{abstract}

%%%%%%%%%%%%%%%%%%%%%%%%%%%%%%%%%%%%%%%%%%%%%%%%%%%%%%%%%%%%%%%%
\begin{figure}[h!]
    \begin{center}
    \includegraphics[width=0.95\columnwidth]{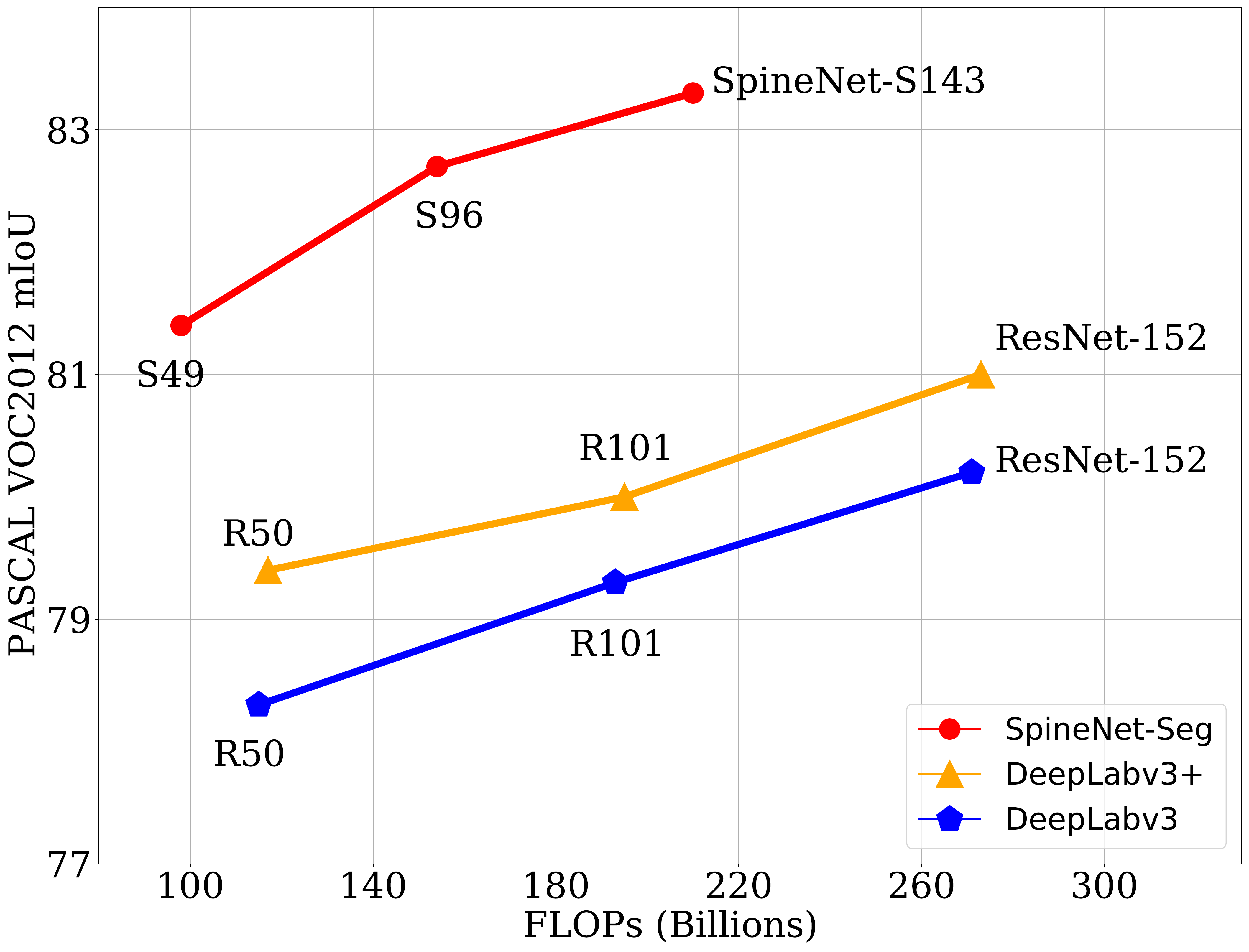}
    \end{center}
    %\vspace{-30mm}
    %\hspace{45mm}\resizebox{.45\columnwidth}{!}{
    %\begin{tabular}[b]{c | c c}
    % & \#FLOPs & mIoU  \\
    %\hline
    %\bf SN-S49 & \bf 44.7B  & \bf 40.6 \\
    %SpineNet-49 & \bf 44.7B  & \bf 40.6 \\
    %\hline
    %\end{tabular}}
    %\vspace{20mm}
    \caption{Performance comparisons of SpineNet-Seg, DeepLabv3 and DeepLabv3+ on the PASCAL VOC2012 \textit{val} set. The proposed SpineNet-Seg models outperform the other two families of models at all model scales. SpineNet-Seg adopts SpineNet-S49/S96/S143 backbones and DeepLabv3 and DeepLabv3+ adopt ResNet-50/101/152 backbones. Controlled experiments and detailed experimental settings can be found in Section~\ref{sec:experiments}.}
    \label{fig:main_results}
\end{figure}
%%%%%%%%%%%%%%%%%%%%%%%%%%%%%%%%%%%%%%%%%%%%%%%%%%%%%%%%%%%%%%%%

\section{Introduction}\label{sec:intro}
Preserving feature resolution and aggregating multi-scale feature information have long been the challenges in achieving better semantic segmentation performance. Convolutional neural networks designed for image-level classification tasks~\cite{alexnet,vgg,inceptionv2,inceptionv3,inceptionv4,xception,resnet,mobilenetv2,mobilenetv3} successively reduce feature resolution by pooling operations and convolutions with strides at different network stages. Such networks only output low-resolution features with strong semantics. \eg ResNet~\cite{resnet} reduces feature resolution to $1/32$ of the input resolution at the end of its $C_5$ stage and only outputs the $C_5$ features. This design is not optimal for semantic segmentation as the pixel-wise classification task benefits from detailed spatial information and aggregation of features from multiple scales.

To solve these problems, researches have proposed better network operations and architecture designs. The dilated convolution operator~\cite{epitomic_conv,Yu2016MultiScaleCA,Chen2015SemanticIS,deeplabv1,deeplabv3,deeplabv3plus} is one of the most popular methods that overcome the challenge of preserving feature resolution. The `convolution with holes' design allows the network to use upsampled convolution kernels to extract abstract semantics without reducing feature resolution. Recently, scale-permuted networks discovered by neural architecture search (NAS)~\cite{spinenet,Du2020EfficientSB} have shown promising results on the task of object detection. Scale permutation for the intermediate building blocks enables the network to capture strong semantics and retain high feature resolution throughout network stages. Cross-scale feature fusion aggregates multi-scale semantics that helps the network to recognize objects at different scales.

%\todo{(xianzhi) summarize the proposed work and results.}
In this work, we first explore the effectiveness of scale-permuted networks on the task of semantic segmentation. We simplify the search space proposed in~\cite{spinenet} and use the backbone of DeepLabv3~\cite{deeplabv3} as the baseline for NAS. The architecture found by NAS improves over the baseline DeepLabv3 model by \textcolor{blue}{+2.06\%} mIoU on the PASCAL VOC2012 benchmark while using less computational resources.
Secondly, we combine dilation convolution with scale-permuted network to further improve semantic segmentation. We delicately design a joint search space for scale permutation, cross-scale connections, block adjustments and block dilation ratios. The final architecture, called SpineNet-Seg-49 (SpineNet-S49), improves mIoU by \textcolor{blue}{+2.47\%} over the baseline on the PASCAL VOC2012 benchmark while using 15\% less computations.
Lastly, we scale and modify the SpineNet-S49 architecture to generate two model families for regular-size semantic segmentation and mobile-size semantic segmentation. In particular, our SpineNet-S143+ model achieves new state-of-the-art performance on Cityscapes at \textbf{83.04\%} and strong performance on PASCAL VOC2012 at \textbf{85.56\%} mIoU, under the settings of single-model single-scale inference without using extra data. Our mobile-size SpineNet-S49- outperforms the MobileNetv3 based DeepLab model by \textcolor{blue}{+2.5\%} while using less computatioinal resources. 

%\todo{(xianzhi) summarize contributions.}
Our contributions are summarized as below:

\begin{itemize}
    \itemsep0em
    \item We prove scale-permuted network improves semantic segmentation.
    \item We propose a novel search space that jointly search for 4 components for semantic segmentation and design a proxy task for NAS.
    \item We outperfrom the baseline DeepLabv3/v3+ models at all model scales by \textcolor{blue}{2-3\%} mIoU on the PASCAL VOC2012 benchmark while using less computations.
    \item We achieve new state-of-the-art on the Cityscapes benchmark at \textbf{83.04\%} mIoU by using \textit{single-model} \textit{single-scale} inference without extra training data.
    \item We provide a family of Mobile SpineNet-Seg models for mobile-size semantic segmentaiton that outperform popular MobileNetv2/v3 based segmentation models. 
\end{itemize}

The remaining contents of the paper are organized as follows. We discuss related works in Section~\ref{sec:related_work}. We describe our search space design and final architectures in Section~\ref{sec:methodology}. The application details for our regular-size and mobile-size segmentation systems are described in~\ref{sec:application}. Our main results and ablation studies are presented in Section~\ref{sec:experiments}. We conclude this work in Section~\ref{sec:conclusion}.

\section{Related Work}\label{sec:related_work}
\paragraph{Semantic segmentation:} Performance of the convolutional neural networks on the task of semantic segmentation has been improved in the recent years by adopting better backbones and improving network designs for semantic segmentation. Since the development of convolutional neural network, researchers have proposed stronger network architectures in better designs and larger scales for the task of image classification, \eg AlexNet~\cite{alexnet}, VGG~\cite{vgg}, Inception~\cite{inceptionv1, inceptionv2, inceptionv3}, ResNet~\cite{resnet}, Xception~\cite{xception}, DenseNet~\cite{densenet}, MobileNet~\cite{mobilenetv2}, Wide ResNet~\cite{zagoruykoK16wideresnet} and ResNeXt~\cite{resnext}. Such networks not only improve image classification, but also transfer to downstream tasks such as object detection, semantic segmentation, depth estimation, \etc. On the other hand, better architecture designs for semantic segmentation have been proposed to preserve object details and to aggregate multi-scale contexts.
~\cite{Shelhamer2017FullyCN,deconvnet,Ronneberger2015UNetCN,Badrinarayanan2017SegNetAD,fpn,Zoph2020RethinkingPA} propose to use an encoder-decoder design to first reduce feature resolution with an encoder to capture deep and coarse semantics then recover spatial resolution via upsampling or deconvolution~\cite{deconv} with a decoder. Shortcut connections can be used between the two components to aggregate multi-scale contexts.
~\cite{deeplabv1,deeplabv3,deeplabv3plus,Zhao2017PyramidSP,He2015SpatialPP} propose to adopt the spatial pyramid pooling module to aggregate context information from local to global at multiple grid scales. 
~\cite{epitomic_conv,Yu2016MultiScaleCA,Chen2015SemanticIS,deeplabv1,deeplabv3,deeplabv3plus,Wu2016BridgingCA,scale-adaptive} advocate to use dilated convolution at certain stages of existing architectures to expand receptive filed of convolution kernels without downsampling. The resulting architectures are able to capture dense semantics without losing resolution.

\paragraph{NAS and search space designs:} NAS automates the design of neural network architecture to find better architectures in a predetermined search space on a target task. Recent architectures discovered by NAS have shown promising results than handcrafted models on vision tasks including image classification~\cite{nasnet,amoebanet,mnasnet,mobilenetv3}, object detection~\cite{nasfpn,spinenet,Du2020EfficientSB,detnet,nasfcos,xu2019autofpn,Liang2020ComputationRF}, semantic segmentation~\cite{autodeeplab,Shaw2019SqueezeNASFN}, \etc. For image classification, typical search space designs include searching for kernel size and filter size of convolutional layers, number of layers per network stage, additional operations such as shortcut connections, attention modules, activation functions, \etc. Recent works have developed customized search space for downstream tasks. For object detection, NAS-FPN~\cite{nasfpn} proposes a search space to search for layer scales and lateral connections for FPN~\cite{fpn}. SpineNet~\cite{spinenet} designs a search space that searches for a new ordering of network blocks for a baseline architecture and cross-scale connections to connect all blocks. CR-NAS~\cite{Liang2020ComputationRF} redistributes computational resources by searching for better block repeating times per network stage for ResNet models. Auto-DeepLab~\cite{autodeeplab} is one of the pioneering works to explore NAS for semantic segmentation. Auto-DeepLab proposes a two-level hierarchical search space that learns operations at block-level and learns block resolutions at network-level with a few handcrafted constraints with respect to common network design choices for semantic segmentation.

\section{Methodology}\label{sec:methodology}
This section starts from introducing our search space for semantic segmentation in Section~\ref{sec:search_space}. The baseline architecture and the computation allocations for NAS are explained in Section~\ref{sec:search_baseline}. The final SpineNet-Seg architecture discovered by NAS and its variants are described in Section~\ref{sec:spinenet_seg_arc} and~\ref{sec:spinenet_seg_mobile}. 

\subsection{Search Space Design}\label{sec:search_space}
The proposed search space consists of 4 components: search a scale permutation for the building blocks of a baseline architecture; search one cross-scale connection for each block; search a level\footnote{Following~\cite{spinenet}, we use ``level" to represent the resolution of a block. $L_i$ indicates a block that has a resolution of $\frac{1}{2^i}$ of the input resolution.} adjustment for each block; search a dilation ratio for the convolution within each block.

\paragraph{Scale permutations and cross-scale connections:} Inspired by~\cite{spinenet}, we define the search space for scale-permutation to be permuting the ordering of intermediate blocks. This results in a search space size of $N$!, where $N$ is the total number of blocks to be permuted. Unlike~\cite{spinenet}, where two cross-scale connections are searched per block, we only search for one long-range connection for each block and simplify the short-range connection to be between each pair of adjacent blocks. This greatly reduces the number of candidates in the search space from $(\prod_{i=m}^{N+m-1} C^i_2)$ to $(\prod_{i=m}^{N+m-1} i)$, where $m$ is the number of initial blocks, while not introducing any performance drop in architecture search.

\paragraph{Block level adjustments:} As the default block level distribution might not be optimal for the target task, we allow each block to search for a level adjustment from a list of integer candidates $\{A_1, A_2, ..., A_a\}$. This results in a search space size of $a^N$.

%ResNet-50+ preserves feature resolution for blocks above $L_4$ and results in significant increase of computations in such high level blocks. In order to keep computations of the architecture candidates in the search space no larger than ResNet-50+, we allow each block to search for a level adjustment within $\{-1, 0\}$, resulting in a search space size of $2^N$. 

\paragraph{Dilation ratios:} Lastly, we introduce the popular dilated convolution operator to the search space. We allow each block to search for one dilation ratio from a list of candidates $\{D_1, D_2, ..., D_d\}$. This results in a search space size of $d^N$.

\subsection{Baseline and Computation Allocations}\label{sec:search_baseline}

%%%%%%%%%%%%%%%%%%%%%%%%%%%%%%%%%%%%%%%%%%%%%%%%%%%%%%
\setlength{\tabcolsep}{4pt}
\begin{table}[h!]
\centering
\begin{tabular}{c| c | c  c}
  \toprule
  Model & Downsample & FLOPs (B) & mIoU  \\
  \midrule
  ResNet-50 & end & 117 & 79.4 \\
  ResNet-S50 & beginning & 85 & 79.2 \\
  \bottomrule
\end{tabular}
\caption{A performance comparison of the original DeepLabv3+ ResNet-50 backbone that downsamples at the end of each stage and our modified ResNet-S50 backbone that downsamples at the beginning of each stage. Results are reported with the DeepLabv3+ system on the PASCAL VOC2012 \textit{val} dataset.}
\label{tab:modified_deeplab_resnet} 
\end{table}
%%%%%%%%%%%%%%%%%%%%%%%%%%%%%%%%%%%%%%%%%%%%%%%%%%%%%%

%%%%%%%%%%%%%%%%%%%%%%%%%%%%%%%%%%%%%%%%%%%%%%%%%%%%%%
\setlength{\tabcolsep}{4pt}
\begin{table}[t!]
\centering
\begin{tabular}{c  |cccccc}
  \toprule
  Block id & BP & CC & LA & DR & FD \\
  \midrule
  $B_0$ & $L_2$ & -- &-- & -- & 64 \\ 
  $B_1$ & $L_2$ & -- & -- & -- & 64 \\
  \midrule
  $B_2$ & $L_3$ & $B_0$ & -1 & 1 & 64\\
  $B_3$ & $L_4$ & $B_1$ & -1 & 2 & 128\\
  $B_4$ & $L_3$ & $B_1$ & 0 & 1 & 128\\
  $B_5$ & $L_3$ & $B_2$ & 0 & 1 & 128\\
  $B_6$ & $L_6$ & $B_3$ & 0 & 1 & 512\\
  $B_7$ & $L_6$ & $B_5$ & -1 & 2 & 512\\
  $B_8$ & $L_7$ & $B_4$ & 0 & 1 & 512\\
  $B_9$ & $L_7$ & $B_6$ & 0 & 4 & 512\\
  $B_{10}$ & $L_5$ & $B_6$ & 0 & 1 & 512\\
  $B_{11}$ & $L_5$ & $B_7$ & 0 & 2 & 512\\
  $B_{12}$ & $L_4$ & $B_8$ & 0 & 4 & 256\\
  $B_{13}$ & $L_4$ & $B_9$ & 0 & 1 & 256\\
  $B_{14}$ & $L_5$ & $B_{11}$ & 0 & 4 & 512 \\
  $B_{15}$ & $L_4$ & $B_{11}$ & 0 & 4 & 256\\
  $B_{16}$ & $L_4$ & $B_{12}$ & 0 & 1 & 256\\
  $B_{17}$ & $L_2$ & $B_{14}$ & 0 & 4 & 64 \\
  $B_{18}$ & $L_7$ & $B_{16}$ & -1 & 2 & 512\\
  $B_{19}$ & $L_6$ & $B_{15}$ & 0 & 1 & 512\\
  $B_{20}$ & $L_4$ & $B_{17}$ & -1 & 4 & 128\\
  \midrule
  $B_{21}$ & - & $B_{0}$ & 0 & 1 & 128\\
  \bottomrule
\end{tabular}
\caption{\textbf{Learned network configurations for the SpineNet-S49 architecture.} We show the detailed configurations for each block for the search space components described in Section~\ref{sec:search_space}. BP: block permutation. CC: cross-scale connection. LA: level adjustment. DR: dilation ratio. FD: feature dimension.}
\label{tab:search_results} 
\end{table}
%%%%%%%%%%%%%%%%%%%%%%%%%%%%%%%%%%%%%%%%%%%%%%%%%%%%%%

Searching for a scale-permuted network starts from a baseline network. In this work, we adopt the ResNet-50 backbone of DeepLabv3~\cite{deeplabv3} with an output stride of 16, and with stage 5 being repeated twice. Unlike DeepLabv3 that proposes to downsample the features at the end of each network stage, we modify the downsampling to happen at the beginning of each stage. This saves ~30\% of the computational cost with negligible loss in performance. A study of the effect of such a modification is shown in Tab.~\ref{tab:modified_deeplab_resnet}. We refer to the modified backbone as ResNet-Seg-50 (ResNet-S50), while we refer to the original DeepLabv3+ backbones as ResNet-50/101/152\footnote{Unless stated otherwise, stage 5 is repeated twice when referring to different ResNet models (e.g. ResNet-50/101/152).}.

%%%%%%%%%%%%%%%%%%%%%%%%%%%%%%%%%%%%%%%%%%%%%%%%%%%%%%
\begin{figure}[t!]
    \begin{center}
    \includegraphics[width=0.7\columnwidth]{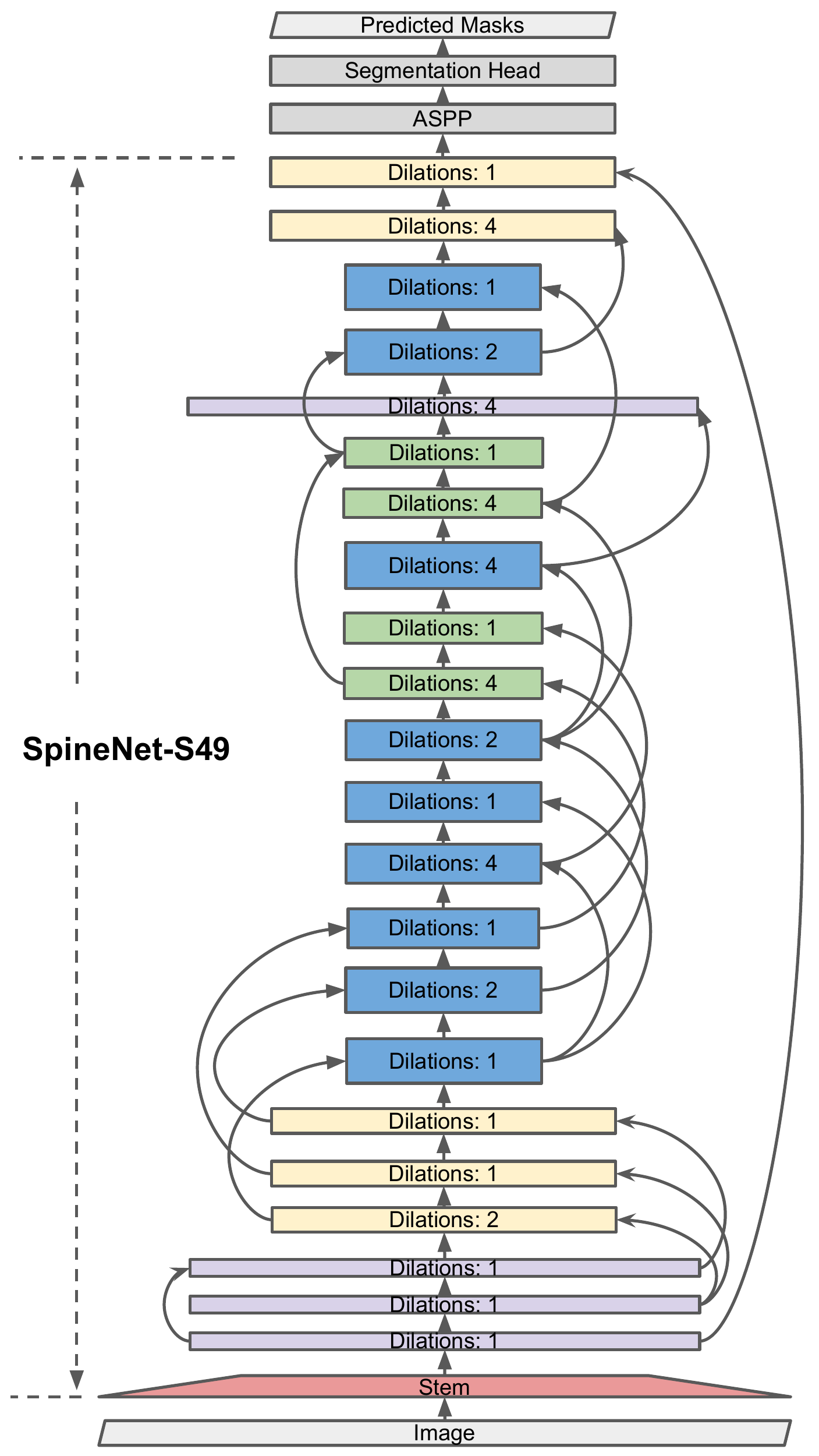}
    \end{center}
    \caption{The final SpineNet-S49 model for semantic segmentation. $\{L_2,L_3,L_4,L_5\}$ blocks are colored in purple, yellow, green and blue, respectively.}
    \label{fig:architecture}
\end{figure}
%\vspace{-10mm}
%%%%%%%%%%%%%%%%%%%%%%%%%%%%%%%%%%%%%%%%%%%%%%%%%%%%%%

ResNet-S50 provides a block allocation of \{3$\times L_2$, 4$\times L_3$, 6$\times L_4$, 9$\times L_5$\} bottleneck blocks. We take two $L_2$ blocks to build an initial network that forms the initial search space for cross-scale connections and reserve one $L_3$ block to construct an output block. For intermediate blocks, we first search for a permutation for the remaining block allocation \{1$\times L_2$, 3$\times L_3$, 6$\times L_4$, 9$\times L_5$\}. Secondly, for each block, we connect it to its immediate previous block and search for one cross-scale connection from the other previous blocks. The same resampling strategy as in~\cite{spinenet} is adopted when merging blocks. Thirdly, we search for one level adjustment within $\{-1, 0\}$. This is because ResNet-S50 keeps feature resolution but increases feature dimension for stage 4 and above. In order to constraint the computation of candidates in the search space to be no larger than the baseline, we only allow each block to either keep or decrease its level (\ie increase feature resolution). Lastly, we search for a dilation ratio within $\{1, 2, 4\}$. The dilation ratio is applied to the 3$\times$3 convolution in each bottleneck block. For the output block, following the common output design for semantic segmentation networks~\cite{deeplabv3,deeplabv3plus,mobilenetv3}, we append a $L_3$ block (\ie output stride 8) at the top and only search for a cross-scale connection and a dilation ratio.

\subsection{SpineNet-Seg Architectures}\label{sec:spinenet_seg_arc}
SpineNet-Seg architectures are searched with a DeepLabv3 system on a semantic segmentation task. The final SpineNet-S49 configuration discovered by NAS are shown in Tab.~\ref{tab:search_results}. Feature dimension $\{32,64,128,256,512\}$ are used for $\{L_1,L_2,L_3,L_4,L_5\}$ blocks, respectively. Based on SpineNet-S49, we construct three larger models, named SpineNet-S96, SpineNet-S143 and SpineNet-S143+, by repeating each block in SpineNet-S49 twice or three times. When repeating one block, we construct replicas of this block and connect them with the original block sequentially without introducing any cross-scale connections. For our largest model SpineNet-S143+, we further uniformly upscale the feature dimension of convolutional layers by $1.3\times$.

We control output stride of the model by changing the size of the output block and its cross-scale connection. We preserve the sizes of the rest of the layers. This is computationally more efficient than changing all layers that are smaller in size than the required output stride as proposed by DeepLabv3~\cite{deeplabv3}. Unless stated, our models have an output stride of 8.

\subsection{Mobile-size SpineNet-Seg Architectures}\label{sec:spinenet_seg_mobile}
Unlike regular semantic segmentation systems that use standard convolution operations, mobile-size systems desire low-computation operations. Inspired by~\cite{mobilenetv2,mobilenetv3,Du2020EfficientSB}, we adopt the inverted bottleneck block that employs depthwise separable convolution~\cite{xception} as its main operator to build mobile-size SpineNet-Seg models.

We construct Mobile SpineNet-S49 and Mobile SpineNet-S49- by replacing all bottleneck blocks with inverted bottleneck blocks. Feature dimension $\{16,24,40,80,112\}$ and expansion ratio 6 are used for $\{L_1,L_2,L_3,L_4,L_5\}$ blocks in Mobile SpineNet-S49, respectively. Mobile SpineNet-S49- uniformly downscales the feature dimension of all convolutional layers by $0.65\times$.

%%%%%%%%%%%%%%%%%%%%%%%%%%%%%%%%%%%%%%%%%%%%
\setlength{\tabcolsep}{4pt}
\begin{table*}[t!]
\centering
\begin{tabular}{c| c | c  c | c}
  \toprule
  Model & Backbone & FLOPs (B) & \#Params (M) & mIoU  \\
  \midrule
  DeepLabv3 & ResNet-50 &  115 & 74 & 78.3 \\
  DeepLabv3+ & ResNet-50 &  117 & 75 & 79.4 \\
  SpineNet-Seg & SpineNet-S49 & 98 & 69 & 81.4 \\
  \midrule
  DeepLabv3 & ResNet-101 & 193 & 93 & 79.3 \\
  DeepLabv3+ & ResNet-101 & 195  & 94 & 79.9 \\
  SpineNet-Seg & SpineNet-S96 & 154  & 116 & 82.6 \\
  \midrule
  DeepLabv3 & ResNet-152 & 271 & 109 & 80.2 \\
  DeepLabv3+ & ResNet-152 & 273  & 110 & 81.0 \\
  SpineNet-Seg & SpineNet-S143 & 210  & 162 & 83.3 \\
  \bottomrule
\end{tabular}
\caption{\textbf{Result comparisons on the PASCAL VOC2012 \textit{val} set.} The proposed SpineNet-Seg models outperform the DeepLabv3 baselines and DeepLabv3+ models at all scales. All models are trained under the same settings.}
\label{tab:pascal_main} 
\end{table*}
%%%%%%%%%%%%%%%%%%%%%%%%%%%%%%%%%%%%%%%%%%%%

\section{Applications}\label{sec:application}
We plug in SpineNet-Seg models as the backbones of the Deeplabv3 system for semantic segmentation. On top of the backbone, we apply an Atrous Spatial Pyramid Pooling (ASPP) module, $n$ convolutions with kernel size 3 and feature dimension 256 followed by batch normalization and activation, and a final classification layer with kernel size 3 to compute pixel-wise predictions. The final architecture of the SpineNet-S49 model is shown in Fig.~\ref{fig:architecture}. Normally, we directly build the final classification layer on top of the ASPP module ($n$ = 0). However, we found that using 2 convolutional layers ($n$ = 2) is essential for stable training when the output stride is 4, and with larger model sizes. In particular, SpineNet-S143+ model is trained with 2 convolutional layers at the head ($n$ = 2), while the rest of the models are trained with $n$ = 0. For mobile-size systems, we replace regular convolutions with depthwise separable convolutions in ASPP and segmentation head.

%%%%%%%%%%%%%%%%%%%%%%%%%%%%%%%%%%%%%%%%%%%%%
\setlength{\tabcolsep}{4pt}
\begin{table}[h!]
\centering
\begin{tabular}{c c c| c}
  \toprule
  EMA & 640$\times$640 & COCO & mIoU  \\
  \midrule
  - & - & - & 81.02\\
  \checkmark & - & - & 81.39 \\
  \checkmark & \checkmark & - & 82.09 \\
  \checkmark & \checkmark & \checkmark & 83.49 \\
  \bottomrule
\end{tabular}
\caption{An ablation study of the training settings. Results are reported with the SpineNet-S49 model on the PASCAL VOC2012 \textit{val} set. EMA: refers to using exponential moving average of the model weights during training. 640x640: using training and evaluation image sizes of 640$\times$640. COCO: Model pretrained on the COCO dataset.}
\label{tab:pascal_ablation} 
\end{table}
%%%%%%%%%%%%%%%%%%%%%%%%%%%%%%%%%%%%%%%%%%%%

\section{Experimental Results}\label{sec:experiments}

We evaluate our Spinenet-Seg models on the PASCAL VOC2012 benchmark~\cite{pascal-voc-2012}, the Cityscapes benchmark~\cite{Cordts2016Cityscapes} and a challenging large-scale Street View dataset.

Unless stated, all experiments (including baselines) use SGD optimizer with momentum of 0.9, cosine learning rate schedule, and exponential moving average (EMA) optimizer with average decay of 0.9998. For pretraining, we use ImageNet~\cite{deng09imagenet} and COCO~\cite{coco} datasets. Unless stated, the results (including DeepLabv3 and DeepLabv3+) reported are with ImageNet pretraining. All results in this section are computed with single scale inference. For fair comparison, we always scale mask predictions to original image sizes to compute mIoU.

%%%%%%%%%%%%%%%%%%%%%%%%%%%%%%%%%%%%%%%%%%%%%%%%%%%%%%%%%%
\setlength{\tabcolsep}{4pt}
\begin{table}[h!]
\centering
\begin{tabular}{c | c | c  c | c}
  \toprule
  Model & Backbone & mIoU  \\
  \midrule
  DFN~\cite{yu2018learning} & ResNet-101 & 80.46 \\
  Auto-Deeplab~\cite{autodeeplab} & Auto-Deeplab-L & 80.75 \\
  GCN~\cite{peng2017large} & ResNet-GCN & 81.00 \\
  DeepLabv3+~\cite{deeplabv3plus} & Xception-65 & 82.45\\
  ExFuse~\cite{zhang2018exfuse} & ResNeXt-131 & 85.40 \\
  \midrule
  SpineNet-Seg & SpineNet-S49 & 83.49 \\
  SpineNet-Seg & SpineNet-S96 & 85.16 \\
  SpineNet-Seg & SpineNet-S143 & 85.64 \\
  \bottomrule
\end{tabular}
\caption{Results on PASCAL VOC2012 \textit{val} set for different SpineNet-Seg models compared to other models. Results in this table include single scale inference with ImageNet, and COCO pretraining.}
\label{tab:pascal_model_scale} 
\end{table}
%%%%%%%%%%%%%%%%%%%%%%%%%%%%%%%%%%%%%%%%%%%%%%%%%%%%%%%%%%

\subsection{Pretraining}
\paragraph{ImageNet pretrain:}
We pretrain the models on ImageNet-1k dataset for 350 epochs with batch size of 4096. We use the following experiment setup for ImageNet pretraining:  cosine learning rate schedule with an initial learning rate of 1.6, regular batch normalization with momentum of 0.99, L2 weight decay of 4e-5, label smoothing of 0.1, we train on random crops of 320x320, we use RandAugment~\cite{randaug} for image augmentations. EMA is not adopted for ImageNet pretraining.

\paragraph{COCO pretrain:}
We use COCO-Things semantic segmentation labels where only annotations for things are used as forground classes, and the rest is used as background. We build the ASPP module such that it matches the ASPP used for the target dataset. We use the following experiment setup for COCO pretraining: we train on image sizes of 512x512 with random horizontal flips, and scale jittering of [0.5, 2.0], batch size of 256, sync batch normalization with momentum of 0.99 and L2 weight decay of 1e-5. We train the models for 64k steps with cosine learning rate schedule with an initial learning rate of 0.08. EMA is not used for COCO pretraining.

%%%%%%%%%%%%%%%%%%%%%%%%%%%%%%%%%%%%%%%%%%%%%%%%%%%%%%%%%
\setlength{\tabcolsep}{4pt}
\begin{table*}[t!]
\centering
\begin{tabular}{c| c | c  c | c}
  \toprule
  Model & Backbone & FLOPs (B) & \#Params (M) & mIoU  \\
  \midrule
  DeepLabv3+ & ResNet-50 & 1092  & 76 & 79.84 \\
  \midrule
  SpineNet-Seg & SpineNet-S49 & 798 & 69 & 81.06 \\
  SpineNet-Seg & SpineNet-S96 & 1272 & 117 & 81.45 \\
  SpineNet-Seg & SpineNet-S143 & 1722 & 164 & 82.11 \\
  SpineNet-Seg$^\dagger$ & SpineNet-S143+ & 2766 & 275 & 83.04 \\
  \bottomrule
\end{tabular}
\caption{\textbf{Result comparisons on the Cityscapes \textit{val} set.} SpineNet-S49 outperforms DeepLabv3+ with a ResNet-50 backbone in both accuracy and speed. SpineNet-S49/S96/S143 and DeepLabv3+ models are trained under the same settings. SpineNet-S143+ marked with $^\dagger$ adopts the best training recipe to achieve best performance.}
\label{tab:cityscapes_main} 
\end{table*}
%%%%%%%%%%%%%%%%%%%%%%%%%%%%%%%%%%%%%%%%%%%%%%%%%%%%%%%%%%
%%%%%%%%%%%%%%%%%%%%%%%%%%%%%%%%%%%%%%%%%%%%%%%%%%%%%%%
\setlength{\tabcolsep}{4pt}
\begin{table*}[h!]
\centering
\begin{tabular}{c| c |  c  | cc}
\toprule
  Model & Backbone & Multi-scale Test & mIoU  \\
  \midrule
  DeepLabv3+~\cite{deeplabv3plus} & Xception-71 & -- & 79.55\\
  MDEQ-XL~\cite{mdeq} & MDEQ & -- & 80.30 \\
  AutoDeeplab~\cite{autodeeplab} & AutoDeeplab-L & -- & 80.33 \\
  RepVGG~\cite{Ding2021RepVGGMV} & RepVGG-B2 & -- & 80.57 \\
  HRNetV2 \cite{hrnet} & HRNetV2-W48 & -- & 81.10  \\
  Panoptic-DeepLab~\cite{Cheng2020PanopticDeepLabAS} & - & -- & 81.50 \\
  HRNetV2 + OCR \cite{hrnet} & HRNetV2-W48 & -- & 81.60  \\
  ResNeSt \cite{zhang2020resnest} & ResNeSt-200 & \checkmark & 82.70 \\
  \midrule
  SpineNet-Seg & SpineNet-S143+ & -- & 83.04 \\
\bottomrule
\end{tabular}
\caption{\textbf{State-of-the-art on the Cityscapes \textit{val} set}. We compare our best model on the Cityscapes \textit{val} set to other models reported in literature. Note that our model uses \textit{single-scale} input for inference and is trained without using extra data.}
\label{tab:cityscapes_second} 
\end{table*}
%%%%%%%%%%%%%%%%%%%%%%%%%%%%%%%%%%%%%%%%%%%%%%%%%%%%%%%
%%%%%%%%%%%%%%%%%%%%%%%%%%%%%%%%%%%%%%%%%%%%%%%%%%%%%%%%%%
\setlength{\tabcolsep}{4pt}
\begin{table}[h!]
\centering
\begin{tabular}{c c c| c}
  \toprule
  EMA & OS=4 & COCO & mIoU  \\
  \midrule
  - & - & - & 81.92\\
  \checkmark & - & - & 82.11 \\
  \checkmark & \checkmark & - & 82.67 \\
  \checkmark & \checkmark & \checkmark & 83.04 \\
  \bottomrule
\end{tabular}
\caption{Cityscapes \textit{val} set results. These results are obtained using single scale inference with no horizontal flipping. OS: refers to the output stride of the SpineNet-Seg model, normally the output stride is 8.}
\label{tab:cityscapes_ablation} 
\end{table}
%%%%%%%%%%%%%%%%%%%%%%%%%%%%%%%%%%%%%%%%%%%%%%%%%%%%%%%%%

\subsection{Results on PASCAL VOC2012}
PASCAL VOC2012~\cite{pascal-voc-2012} is a semantic segmentation dataset with 20 forground classes and 1 background class. For training, we use an augmented version of the dataset~\cite{extra_seg_data} with extra annotations of 10582 images (trainaug).
The default training setup uses training image sizes of 512x512 with scale jittering of [0.5, 2.0] and random horizontal image flipping. We use batch size of 32, and sync batch normalization with momentum of 0.9997.  We use dilation rates of 12, 24 and 36 to build ASPP. We train experiments for 20k steps.

Tab.~\ref{tab:pascal_main} and Fig.~\ref{fig:main_results} shows performance comparisons of our SpineNet-Seg models \vs DeepLabv3 and DeepLabv3+ with counterpart ResNet backbones. All models are trained using the same experiment setup. Our results show consistent +3\% and +2\% improvements in mIoU across all model scales compared to DeepLabv3 and DeepLabv3+ models. Specifically, SpineNet-S49, SpineNet-S96, and SpineNet-S143 show improvements of +2\%, +2.66\%, and +2.29\% in mIoU compared to DeepLabv3+ with ResNet-50, ResNet-101, and ResNet-152 backbones respectively. While having significant gain over DeepLab models, SpineNet models are less computationally expensive than their Deeplab ResNet counterparts.

Tab.~\ref{tab:pascal_ablation} studies the effect of using different training setups on the PASCAL VOC2012 validation set. We found that using EMA of model weights improved mIoU by 0.37\%. We also increase the training and evaluation image sizes to 640$\times$640 and resize the prediction masks to its original image sizes for fair comparisons. Using image sizes of 640$\times$640 shows an improvement of 0.7\%. Finally, pretraining on COCO dataset shows an improvement of 1.4\% in mIoU. As a result, the mIoU on PASCAL validation set of our SpineNet-S49 model achieves 83.49\% mIoU.

Tab.~\ref{tab:pascal_model_scale} summarizes the effect of scaling up the model size by using different block repeats of 1, 2, and 3 (SpineNet-S49, SpineNet-S96, and SpineNet-S143 respectively). In these experiments, we used the best training setup in Tab.~\ref{tab:pascal_ablation}. SpineNet-S96 improves the mIoU by 1.67\% on PASCAL Validation set, and SpineNet-S143 improves the mIoU by another 0.48\%. Our best model using single scale inference achieves 85.64\% mIoU on Pascal VOC validation set. We also compare our best models with previous work in Tab.~\ref{tab:pascal_model_scale}.

\subsection{Results on Cityscapes}
Cityscapes~\cite{Cordts2016Cityscapes} contains high quality pixel-level annotations of 5000 images (2975, 500, and 1525 for train, validation, and test splits respectively). It also contains 20000 coarsely annotated images for training. In our experiments, we only used the high quality pixel-level annotation train split for training, and evaluate on the validation split. Following~\cite{Cordts2016Cityscapes}, we train and evaluate on 19 semantic labels and ignore the void label.

We train on crops of 512x1024 with scale jittering of [0.5, 2.0] and random horizontal image flipping. We use batch size of 64, and sync batch normalization with momentum of 0.99. We use dilation rates of 12, 24, 36, and 72 to build ASPP. We train each experiment for 100k iterations.

Tab.~\ref{tab:cityscapes_main} compares SpineNet-S49 to DeepLabv3+ with ResNet-50 backbone. Both models are trained using the same training setup including ImageNet pretraining, batch size, and using EMA of model weights. Our SpineNet-S49 model shows an improvement of +1.22\% in mIoU compared to its DeepLabv3+ ResNet-50 model counterpart. In the same table, we show the effect of scaling up the model using block repeats of 1, 2, and 3 (SpineNet-S49, SpineNet-S96, and SpienNet-S143 respectively). SpineNet-S96 model improves the performance by 0.39\%, while SpineNet-S143 further improves the mIoU by another 0.66\%.

Tab.~\ref{tab:cityscapes_ablation} shows the effect of using different training setup on the Cityscapes validation set. We used SpineNet-S143 model (block repeats of 3 and output stride of 8) as the baseline for the ablation study. We found that using exponential moving average of model weights improved mIoU by 0.19\%. Moreover, we adopt the largest backbone SpineNet-S143+ and change the output stride of the model to 4. SpineNet-S143+ improves the mIoU by 0.56\% on Cityscapes validation set. Finally, we pretrain SpineNet-S143+ on COCO and finetune on Cityscapes which further improves performance by 0.37\%. As shown in Tab.~\ref{tab:cityscapes_second}, our best model at 83.04\% mIoU on Cityscapes validation set achieves the new state-of-the-art when using single scale inference.

\subsubsection{Mobile SpineNet-Seg Results}
%%%%%%%%%%%%%%%%%%%%%%%%%%%%%%%%%%%%%%%%%%%%%%%%%%%%%%
\setlength{\tabcolsep}{4pt}
\begin{table}[h!]
\centering
\begin{tabular}{c| c | c  c | c}
  \toprule
  Model & Params (M) & mIoU  \\
  \midrule
  MobileNetV3 \cite{mobilenetv3} & 3.60 & 72.64\\
  \midrule
  Mobile SpineNet-S49- & 3.15 & 75.18 \\
  Mobile SpineNet-S49 & 4.40 & 77.41 \\
  \bottomrule
\end{tabular}
\caption{\textbf{Mobile SpineNet-Seg results on Cityscapes \textit{val} dataset.} We compare our models to MobileNetV3 version.}
\label{tab:cityscapes_mobilenet} 
\end{table}
%%%%%%%%%%%%%%%%%%%%%%%%%%%%%%%%%%%%%%%%%%%%%%%%%%%%%%
%As explained in Sec. \ref{sec:mobilespinenet}, Mobile SpineNet-49+ are constructed by replacing the Bottleneck block by Inverted Bottleneck blocks. We also replaced regular convolutions by depthwise separable convolutions in ASPP module.
We follow the training setup for SpineNet-Seg models to train two mobile-size models: Mobile SpineNet-S49 and Mobile SpineNet-S49-. As shown in Tab.~\ref{tab:cityscapes_mobilenet}, Mobile SpineNet-Seg models achieve significantly better mIoU in speed and accuracy compared to MobileNetV3 model.

\subsection{Results on Street View}
%%%%%%%%%%%%%%%%%%%%%%%%%%%%%%%%%%%%%%%%%%%%%%%%%%%%%%
% From slide deck: https://docs.google.com/presentation/d/1F4rIjmsO8tejLrA00PvOAi-G_tADR_XbvibpJSOK-n0/edit?usp=sharing&resourcekey=0-Osl-J_GVXuAGpF3M2Q0-dg
% CBDB image id: geLxxAuLodEAAAQ5vCH4Og
% CBDB run id: 20171011_193114_L15124
% More images can be selected this way: https://txui.corp.google.com/get?cmd=from+%2Fcns%2Fue-d%2Fhome%2Fyinx%2Fscratch%2Frs%3D6.4%2Fgeo_streetview%2Fprod%2F2021-02-24%2FCAR%2Fval%401000+select+image%2Fencoded%2Cimage%2Fsegmentation%2Fclass%2Fvisualization%2Fencoded%2Cimage%2Fid%2Cimage%2Fmetadata%2Frun_id%2Cimage%2Fmetadata%2Fregion+where+image%2Fmetadata%2Fregion%3D%27sanfrancisco%27++limit+100
\begin{figure}[h!]
    \centering
    \includegraphics[width=0.9\linewidth]{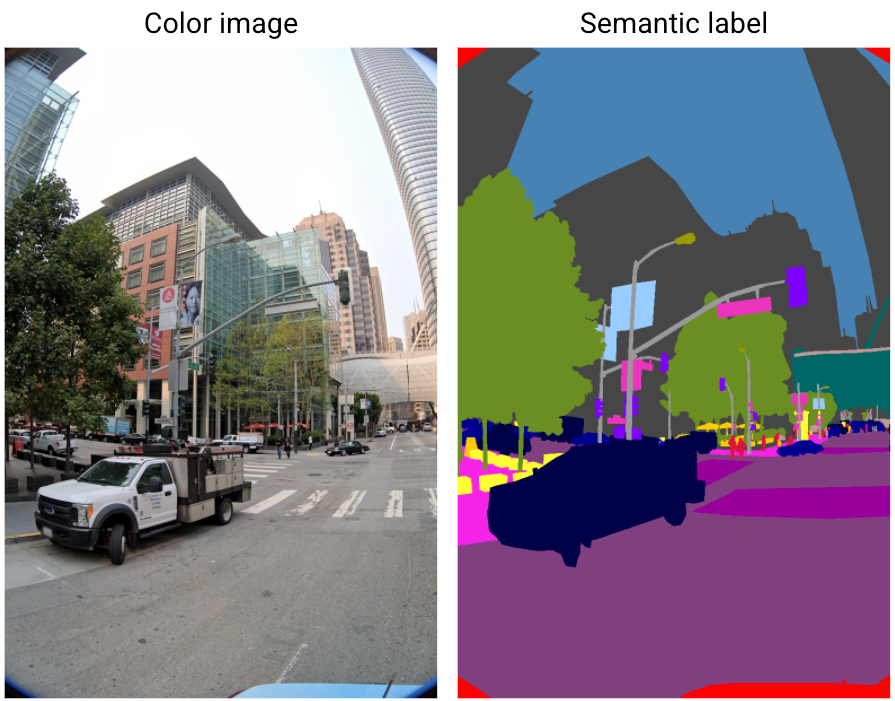}
    \caption{Example Street View images with semantic labels}
    \label{fig:streetview_example}
\end{figure}
%%%%%%%%%%%%%%%%%%%%%%%%%%%%%%%%%%%%%%%%%%%%%%%%%%%%%%

We further evaluate SpineNet-Seg on a challenging dataset from Street View images. The dataset contains 57k train images and 13k test images, with 44 semantic categories on typical street scenes, e.g., building, sidewalk, traffic sign, and cars. In addition, the dataset is collected across 6 continents, 39 countries and 80+ cities worldwide under diverse conditions. Fig.~\ref{fig:streetview_example} shows an typical example of the dataset. Given the complexity, size and geo-diversity, we believe this is a practical stress test for the proposed model.

For controlled experiment, we compare our SpineNet-S143+ to ResNet-152 using the same experiment settings. We use train and eval image sizes of 1152x768 with scale jittering of [0.5, 2.0] and random horizontal image flipping. We use batch size of 128, and sync batch normalization with momentum of 0.99. We use dilation rates of 12, 24, 36, and 72 to build ASPP. We train each experiment for 100k iterations. We also compare to DeepLabv3+ with Xception65 backbone when trained according to experiment setting suggested in~\cite{deeplabv3plus}. All experiments are trained and evaluated on the same image sizes.

Tab.~\ref{tab:results_sv} compares SpineNet-Seg to DeepLabv3+ models. SpineNet model improves the mIoU by 1.64\% compared to DeepLabv3+ with ResNet-152 backbone. We also observe 2.46\% improvements on mIOU compared to DeepLabv3+ with Xception-65 backbone. The results support our claim that SpineNet-Seg outperform DeepLabv3+ models in challenging real world scenarios.

%%%%%%%%%%%%%%%%%%%%%%%%%%%%%%%%%%%%%%%%%%%%%%%%%%%%%%
\setlength{\tabcolsep}{4pt}
\begin{table}[h!]
\centering
\begin{tabular}{c| c | c  c | c}
  \toprule
  Model & Backbone & mIoU  \\
  \midrule
  DeepLabv3+ & Xception-65 & 57.06 \\
  DeepLabv3+ & ResNet-152 & 57.88 \\
  SpineNet-Seg & SpineNet-S143+ & 59.52 \\
  \bottomrule
\end{tabular}
\caption{Results on Street View Dataset.}
\label{tab:results_sv} 
\end{table}
%%%%%%%%%%%%%%%%%%%%%%%%%%%%%%%%%%%%%%%%%%%%%%%%%%%%%%

\subsection{NAS Experiments}
\subsubsection{NAS implementation details}
We run NAS for 10k trials, and we evaluate the best 10 architectures on PASCAL VOC2012. Due to the large number of trials, we design a proxy search task to quickly evaluate the architecture candidates. For all search experiments, we uniformly downscale the feature dimension of convolutional layers of the candidate models by $0.5\times$ and use image sizes 384$\times$384 for training and evaluation. We run training for 30k steps with batch size of 64 and collect the final evaluation mIoU as the reward for the controller~\cite{nas}.

%%%%%%%%%%%%%%%%%%%%%%%%%%%%%%%%%%%%%%%%%%%%%%%%%%%%%%
\setlength{\tabcolsep}{4pt}
\begin{table}[h!]
\centering
\begin{tabular}{c| c | ccc }
  \toprule
  Model &  Baseline & R-S50$^\dagger$ & R-S50 & R-S101$^\dagger$  \\
  \midrule
  mIoU Gain (\%) & +0 & +0.79 & +2.47 & +2.02\\
  \bottomrule
\end{tabular}
\caption{\textbf{Effectiveness of different search baselines.} We show mIoU gain on PASCAL VOC2012 \texttt{val} set when using different architectures, ResNet-S50 and ResNet-S101, as the baseslines for NAS. Backbones marked with $^\dagger$ indicate stage 5 not repeated.}
\label{tab:search_models} 
\end{table}
%%%%%%%%%%%%%%%%%%%%%%%%%%%%%%%%%%%%%%%%%%%%%%%%%%%%%%
%%%%%%%%%%%%%%%%%%%%%%%%%%%%%%%%%%%%%%%%%%%%%%%%%%%%%%
\setlength{\tabcolsep}{4pt}
\begin{table}[h!]
\centering
\begin{tabular}{c| c |c c}
  \toprule
  Output Stride & Baseline & SP & SP+DR  \\
  \midrule
  mIoU Gain & +0 & +2.15 & +2.47 \\
  \bottomrule
\end{tabular}
\caption{\textbf{An ablation of searching for scale-permuted network and dilation ratios.} We show mIoU gain on PASCAL VOC2012 \texttt{val} set when searching for scale-permuted network or jointly searching for scale-permuted network and dilation ratios. SP: scale-permuted network. DR: dilation ratio.}
\label{tab:search_perm_dils} 
\end{table}
%%%%%%%%%%%%%%%%%%%%%%%%%%%%%%%%%%%%%%%%%%%%%%%%%%%%%%
%%%%%%%%%%%%%%%%%%%%%%%%%%%%%%%%%%%%%%%%%%%%%%%%%%%%%%
\setlength{\tabcolsep}{4pt}
\begin{table}[h!]
\centering
\begin{tabular}{c| c |c c c}
  \toprule
  Output Stride & Baseline & 4 & 8 & 16  \\
  \midrule
  mIoU gain & +0 & +0.67 & +2.15 & +1.47 \\
  \bottomrule
\end{tabular}
\caption{\textbf{Impact of different output strides.} This table shows mIoU gain on PASCAL VOC2012 \texttt{val} set when searching architectures with different output strides.}
\label{tab:search_os} 
\end{table}
%%%%%%%%%%%%%%%%%%%%%%%%%%%%%%%%%%%%%%%%%%%%%%%%%%%%%%
%%%%%%%%%%%%%%%%%%%%%%%%%%%%%%%%%%%%%%%%%%%%%%%%%%%%%%
\setlength{\tabcolsep}{4pt}
\begin{table}[h!]
\centering
\begin{tabular}{c| c |c c}
  \toprule
  Search Dataset & Baseline & COCO Things & Stuff+Things  \\
  \midrule
  mIoU Gain & +0 & +2.47 & +2.06 \\
  \bottomrule
\end{tabular}
\caption{\textbf{A study on the search dataset.} We show mIoU gain on PASCAL VOC2012 \texttt{val} set when using COCO Things or COCO Stuff+Things for NAS.}
\label{tab:search_coco} 
\end{table}
%%%%%%%%%%%%%%%%%%%%%%%%%%%%%%%%%%%%%%%%%%%%%%%%%%%%%%

\subsubsection{Ablation studies of the search designs}
First, we experiment with different baseline architectures to search from. Inspired by DeepLabv3, we consider three architectures, ResNet-S50$^\dagger$, ResNet-S50 and ResNet-S101$^\dagger$, where $^\dagger$ indicates absence of stages 6 and 7. Tab.~\ref{tab:search_models} shows that searching from ResNet-S50 yields best improvements with +2.47\% in mIoU gain over the baseline.
We also study the effect of searching for scale-permuted network and dilation ratios in Tab.~\ref{tab:search_perm_dils}. We found that searching for scale-permuted networks yields +2.15\% in mIoU gain. Further searching for dilation ratios improves the mIoU by +0.32\%.
Finally, we study the effect of fixing the output stride to 4, 8, or 16 during search in Tab.~\ref{tab:search_os}, while also changing the feature dimension of the output block accordingly such that the model size is preserved among the search jobs. We found that output stride of 8 yields the best gain of +2.15\% in mIoU.

\subsubsection{A study on search dataset}
Search dataset is important since the evaluation signals influence the quality of the searched architectures. For instance, performance on the PASCAL VOC2012 dataset depends heavily on the quality of the pretrained checkpoint, hence it is difficult to use for NAS that trains proxty tasks from scratch. We decide to use COCO dataset since it is diverse, and unlike Cityscapes it has significantly smaller images. Training proxy tasks from scratch using the COCO dataset usually converges, and eval signals can be used as stable rewards to update the NAS controller. We study the effect of using COCO Things annotations and COCO Things+Stuff annotations. Tab.~\ref{tab:search_coco} shows that COCO Things achieves better performance (+2.47\% mIoU gain) compared to COCO Things+Stuff (+2.06\%).

\section{Conclusion}\label{sec:conclusion}
In this work, we evaluated the effectiveness of scale-permuted architectures on the task of semantic segmentation, a vision task that benefits from high feature resolution and multi-scale feature fusion. We proposed a new search space that simplifies the SpineNet search space and introduces new search components for semantic segmentation and learned SpineNet-S49 architecture by NAS with a carefully designed proxy task. We further construct two families of models based on SpineNet-S49: SpnieNet-Seg models and Mobile SpineNet-Seg models. SpineNet-Seg models outperform the popular DeepLabv3/v3+ models on the PASCAL VOC2012 benchmark and a challenging Street View data and achieve state-of-the-art performance on the Cityscapes benchmark. Mobile SpineNet-Seg models achieve new state-of-the-art performance on mobile-size semantic segmentation, surpassing popular mobile segmentation systems such as MobileNetV2/V3. We expect scale-permuted network with task-specific designs to benefit more computer vision tasks in the future.

\clearpage

{\small
\bibliographystyle{ieee_fullname}
\bibliography{egbib}

\begin{thebibliography}{10}\itemsep=-1pt

\bibitem{Badrinarayanan2017SegNetAD}
Vijay Badrinarayanan, Alex Kendall, and R. Cipolla.
\newblock Segnet: A deep convolutional encoder-decoder architecture for image
  segmentation.
\newblock {\em IEEE Transactions on Pattern Analysis and Machine Intelligence},
  39:2481--2495, 2017.

\bibitem{mdeq}
Shaojie Bai, Vladlen Koltun, and J.~Zico Kolter.
\newblock Multiscale deep equilibrium models.
\newblock In H. Larochelle, M. Ranzato, R. Hadsell, M.~F. Balcan, and H. Lin,
  editors, {\em Advances in Neural Information Processing Systems}, volume~33,
  pages 5238--5250. Curran Associates, Inc., 2020.

\bibitem{Chen2015SemanticIS}
Liang-Chieh Chen, G. Papandreou, I. Kokkinos, Kevin Murphy, and A. Yuille.
\newblock Semantic image segmentation with deep convolutional nets and fully
  connected crfs.
\newblock {\em CoRR}, abs/1412.7062, 2015.

\bibitem{deeplabv1}
Liang-Chieh Chen, G. Papandreou, I. Kokkinos, Kevin Murphy, and A. Yuille.
\newblock Deeplab: Semantic image segmentation with deep convolutional nets,
  atrous convolution, and fully connected crfs.
\newblock {\em IEEE Transactions on Pattern Analysis and Machine Intelligence},
  40:834--848, 2018.

\bibitem{deeplabv3}
Liang-Chieh Chen, G. Papandreou, Florian Schroff, and H. Adam.
\newblock Rethinking atrous convolution for semantic image segmentation.
\newblock {\em ArXiv}, abs/1706.05587, 2017.

\bibitem{deeplabv3plus}
Liang-Chieh Chen, Y. Zhu, G. Papandreou, Florian Schroff, and H. Adam.
\newblock Encoder-decoder with atrous separable convolution for semantic image
  segmentation.
\newblock {\em ArXiv}, abs/1802.02611, 2018.

\bibitem{Cheng2020PanopticDeepLabAS}
Bowen Cheng, Maxwell~D. Collins, Y. Zhu, T. Liu, T. Huang, H. Adam, and
  Liang-Chieh Chen.
\newblock Panoptic-deeplab: A simple, strong, and fast baseline for bottom-up
  panoptic segmentation.
\newblock {\em 2020 IEEE/CVF Conference on Computer Vision and Pattern
  Recognition (CVPR)}, pages 12472--12482, 2020.

\bibitem{xception}
Fran{\c{c}}ois Chollet.
\newblock Xception: Deep learning with depthwise separable convolutions.
\newblock In {\em CVPR}, 2017.

\bibitem{Cordts2016Cityscapes}
Marius Cordts, Mohamed Omran, Sebastian Ramos, Timo Rehfeld, Markus Enzweiler,
  Rodrigo Benenson, Uwe Franke, Stefan Roth, and Bernt Schiele.
\newblock The cityscapes dataset for semantic urban scene understanding.
\newblock In {\em Proc. of the IEEE Conference on Computer Vision and Pattern
  Recognition (CVPR)}, 2016.

\bibitem{randaug}
Ekin~D. Cubuk, Barret Zoph, Jonathon Shlens, and Quoc~V. Le.
\newblock Randaugment: Practical automated data augmentation with a reduced
  search space, 2019.

\bibitem{deng09imagenet}
Jia Deng, Wei Dong, Richard Socher, Li-Jia Li, Kai Li, and Li Fei-Fei.
\newblock Imagenet: A large-scale hierarchical image database.
\newblock In {\em CVPR}, 2009.

\bibitem{Ding2021RepVGGMV}
Xiaohan Ding, X. Zhang, Ningning Ma, J. Han, G. Ding, and Jian Sun.
\newblock Repvgg: Making vgg-style convnets great again.
\newblock {\em ArXiv}, abs/2101.03697, 2021.

\bibitem{Du2020EfficientSB}
Xianzhi Du, Tsung-Yi Lin, Pengchong Jin, Yin Cui, M. Tan, Quoc~V. Le, and
  Xiaodan Song.
\newblock Efficient scale-permuted backbone with learned resource distribution.
\newblock {\em ArXiv}, abs/2010.11426, 2020.

\bibitem{spinenet}
Xianzhi Du, Tsung-Yi Lin, Pengchong Jin, Golnaz Ghiasi, Mingxing Tan, Yin Cui,
  Quoc~V. Le, and Xiaodan Song.
\newblock Spinenet: Learning scale-permuted backbone for recognition and
  localization.
\newblock In {\em Proceedings of the IEEE/CVF Conference on Computer Vision and
  Pattern Recognition (CVPR)}, June 2020.

\bibitem{pascal-voc-2012}
M. Everingham, L. Van~Gool, C.~K.~I. Williams, J. Winn, and A. Zisserman.
\newblock The {PASCAL} {V}isual {O}bject {C}lasses {C}hallenge 2012 {(VOC2012)}
  {R}esults.
\newblock
  http://www.pascal-network.org/challenges/VOC/voc2012/workshop/index.html.

\bibitem{nasfpn}
Golnaz Ghiasi, Tsung-Yi Lin, and Quoc~V Le.
\newblock Nas-fpn: Learning scalable feature pyramid architecture for object
  detection.
\newblock In {\em CVPR}, 2019.

\bibitem{extra_seg_data}
Bharath Hariharan, Pablo Arbelaez, Lubomir Bourdev, Subhransu Maji, and
  Jitendra Malik.
\newblock Semantic contours from inverse detectors.
\newblock In {\em International Conference on Computer Vision (ICCV)}, 2011.

\bibitem{He2015SpatialPP}
Kaiming He, X. Zhang, Shaoqing Ren, and Jian Sun.
\newblock Spatial pyramid pooling in deep convolutional networks for visual
  recognition.
\newblock {\em IEEE Transactions on Pattern Analysis and Machine Intelligence},
  37:1904--1916, 2015.

\bibitem{resnet}
Kaiming He, Xiangyu Zhang, Shaoqing Ren, and Jian Sun.
\newblock Deep residual learning for image recognition.
\newblock In {\em CVPR}, 2016.

\bibitem{mobilenetv3}
Andrew Howard, Mark Sandler, Grace Chu, Liang-Chieh Chen, Bo Chen, Mingxing
  Tan, Weijun Wang, Yukun Zhu, Ruoming Pang, Vijay Vasudevan, et~al.
\newblock Searching for mobilenetv3.
\newblock In {\em ICCV}, 2019.

\bibitem{densenet}
Gao Huang, Zhuang Liu, Laurens Van Der~Maaten, and Kilian~Q Weinberger.
\newblock Densely connected convolutional networks.
\newblock In {\em CVPR}, 2017.

\bibitem{alexnet}
Alex Krizhevsky, Ilya Sutskever, and Geoffrey~E Hinton.
\newblock Imagenet classification with deep convolutional neural networks.
\newblock In {\em Advances in Neural Information Processing Systems}, 2012.

\bibitem{detnet}
Zeming Li, Chao Peng, Gang Yu, Xiangyu Zhang, Yangdong Deng, and Jian Sun.
\newblock Detnet: Design backbone for object detection.
\newblock In {\em ECCV}, 2018.

\bibitem{Liang2020ComputationRF}
Feng Liang, Chen Lin, Ronghao Guo, Ming Sun, Wei Wu, J. Yan, and Wanli Ouyang.
\newblock Computation reallocation for object detection.
\newblock {\em ArXiv}, abs/1912.11234, 2020.

\bibitem{fpn}
Tsung-Yi Lin, Piotr Doll{\'a}r, Ross Girshick, Kaiming He, Bharath Hariharan,
  and Serge Belongie.
\newblock Feature pyramid networks for object detection.
\newblock In {\em CVPR}, 2017.

\bibitem{coco}
Tsung-Yi Lin, Michael Maire, Serge Belongie, James Hays, Pietro Perona, Deva
  Ramanan, Piotr Doll{\'a}r, and C~Lawrence Zitnick.
\newblock Microsoft coco: Common objects in context.
\newblock In {\em ECCV}, 2014.

\bibitem{autodeeplab}
Chenxi Liu, Liang-Chieh Chen, Florian Schroff, Hartwig Adam, Wei Hua, Alan~L
  Yuille, and Li Fei-Fei.
\newblock Auto-deeplab: Hierarchical neural architecture search for semantic
  image segmentation.
\newblock In {\em CVPR}, 2019.

\bibitem{deconvnet}
Hyeonwoo Noh, Seunghoon Hong, and B. Han.
\newblock Learning deconvolution network for semantic segmentation.
\newblock {\em 2015 IEEE International Conference on Computer Vision (ICCV)},
  pages 1520--1528, 2015.

\bibitem{epitomic_conv}
G. {Papandreou}, I. {Kokkinos}, and P. {Savalle}.
\newblock Modeling local and global deformations in deep learning: Epitomic
  convolution, multiple instance learning, and sliding window detection.
\newblock In {\em 2015 IEEE Conference on Computer Vision and Pattern
  Recognition (CVPR)}, pages 390--399, 2015.

\bibitem{peng2017large}
Chao Peng, Xiangyu Zhang, Gang Yu, Guiming Luo, and Jian Sun.
\newblock Large kernel matters--improve semantic segmentation by global
  convolutional network.
\newblock In {\em Proceedings of the IEEE conference on computer vision and
  pattern recognition}, pages 4353--4361, 2017.

\bibitem{amoebanet}
Esteban Real, Alok Aggarwal, Yanping Huang, and Quoc~V Le.
\newblock Regularized evolution for image classifier architecture search.
\newblock In {\em AAAI}, 2019.

\bibitem{Ronneberger2015UNetCN}
O. Ronneberger, P. Fischer, and T. Brox.
\newblock U-net: Convolutional networks for biomedical image segmentation.
\newblock {\em ArXiv}, abs/1505.04597, 2015.

\bibitem{mobilenetv2}
Mark Sandler, Andrew Howard, Menglong Zhu, Andrey Zhmoginov, and Liang-Chieh
  Chen.
\newblock Mobilenetv2: Inverted residuals and linear bottlenecks.
\newblock In {\em CVPR}, 2018.

\bibitem{Shaw2019SqueezeNASFN}
Albert~Eaton Shaw, D. Hunter, Forrest~N. Iandola, and S. Sidhu.
\newblock Squeezenas: Fast neural architecture search for faster semantic
  segmentation.
\newblock {\em 2019 IEEE/CVF International Conference on Computer Vision
  Workshop (ICCVW)}, pages 2014--2024, 2019.

\bibitem{Shelhamer2017FullyCN}
Evan Shelhamer, J. Long, and Trevor Darrell.
\newblock Fully convolutional networks for semantic segmentation.
\newblock {\em IEEE Transactions on Pattern Analysis and Machine Intelligence},
  39:640--651, 2017.

\bibitem{vgg}
Karen Simonyan and Andrew Zisserman.
\newblock Very deep convolutional networks for large-scale image recognition.
\newblock In {\em ICLR}, 2015.

\bibitem{inceptionv4}
Christian Szegedy, Sergey Ioffe, Vincent Vanhoucke, and Alexander~A Alemi.
\newblock Inception-v4, inception-resnet and the impact of residual connections
  on learning.
\newblock In {\em AAAI}, 2017.

\bibitem{inceptionv1}
Christian Szegedy, Wei Liu, Yangqing Jia, Pierre Sermanet, Scott Reed, Dragomir
  Anguelov, Dumitru Erhan, Vincent Vanhoucke, and Andrew Rabinovich.
\newblock Going deeper with convolutions.
\newblock In {\em CVPR}, 2015.

\bibitem{inceptionv2}
Christian Szegedy, Vincent Vanhoucke, Sergey Ioffe, Jonathon Shlens, and
  Zbigniew Wojna.
\newblock Rethinking the inception architecture for computer vision.
\newblock {\em CoRR}, abs/1512.00567, 2015.

\bibitem{inceptionv3}
Christian Szegedy, Vincent Vanhoucke, Sergey Ioffe, Jon Shlens, and Zbigniew
  Wojna.
\newblock Rethinking the inception architecture for computer vision.
\newblock In {\em CVPR}, 2016.

\bibitem{mnasnet}
Mingxing Tan, Bo Chen, Ruoming Pang, Vijay Vasudevan, Mark Sandler, Andrew
  Howard, and Quoc~V Le.
\newblock Mnasnet: Platform-aware neural architecture search for mobile.
\newblock In {\em CVPR}, 2019.

\bibitem{hrnet}
Jingdong Wang, Ke Sun, Tianheng Cheng, Borui Jiang, Chaorui Deng, Yang Zhao,
  Dong Liu, Yadong Mu, Mingkui Tan, Xinggang Wang, et~al.
\newblock Deep high-resolution representation learning for visual recognition.
\newblock {\em PAMI}, 2020.

\bibitem{nasfcos}
Ning Wang, Yang Gao, Hao Chen, Peng Wang, Zhi Tian, and Chunhua Shen.
\newblock {NAS-FCOS}: Fast neural architecture search for object detection.
\newblock {\em arXiv preprint arXiv:1906.04423}, 2019.

\bibitem{Wu2016BridgingCA}
Zifeng Wu, Chunhua Shen, and A.~V.~D. Hengel.
\newblock Bridging category-level and instance-level semantic image
  segmentation.
\newblock {\em ArXiv}, abs/1605.06885, 2016.

\bibitem{resnext}
Saining Xie, Ross Girshick, Piotr Doll{\'a}r, Zhuowen Tu, and Kaiming He.
\newblock Aggregated residual transformations for deep neural networks.
\newblock In {\em CVPR}, 2017.

\bibitem{xu2019autofpn}
Hang Xu, Lewei Yao, Wei Zhang, Xiaodan Liang, and Zhenguo Li.
\newblock Auto-fpn: Automatic network architecture adaptation for object
  detection beyond classification.
\newblock In {\em ICCV}, 2019.

\bibitem{yu2018learning}
Changqian Yu, Jingbo Wang, Chao Peng, Changxin Gao, Gang Yu, and Nong Sang.
\newblock Learning a discriminative feature network for semantic segmentation.
\newblock In {\em Proceedings of the IEEE conference on computer vision and
  pattern recognition}, pages 1857--1866, 2018.

\bibitem{Yu2016MultiScaleCA}
F. Yu and V. Koltun.
\newblock Multi-scale context aggregation by dilated convolutions.
\newblock {\em CoRR}, abs/1511.07122, 2016.

\bibitem{zagoruykoK16wideresnet}
Sergey Zagoruyko and Nikos Komodakis.
\newblock Wide residual networks.
\newblock In {\em BMVC}, 2016.

\bibitem{deconv}
M.~D. {Zeiler}, G.~W. {Taylor}, and R. {Fergus}.
\newblock Adaptive deconvolutional networks for mid and high level feature
  learning.
\newblock In {\em 2011 International Conference on Computer Vision}, pages
  2018--2025, 2011.

\bibitem{zhang2020resnest}
Hang Zhang, Chongruo Wu, Zhongyue Zhang, Yi Zhu, Zhi Zhang, Haibin Lin, Yue
  Sun, Tong He, Jonas Mueller, R. Manmatha, Mu Li, and Alexander Smola.
\newblock Resnest: Split-attention networks, 2020.

\bibitem{scale-adaptive}
R. {Zhang}, S. {Tang}, Y. {Zhang}, J. {Li}, and S. {Yan}.
\newblock Scale-adaptive convolutions for scene parsing.
\newblock In {\em 2017 IEEE International Conference on Computer Vision
  (ICCV)}, pages 2050--2058, 2017.

\bibitem{zhang2018exfuse}
Zhenli Zhang, Xiangyu Zhang, Chao Peng, Xiangyang Xue, and Jian Sun.
\newblock Exfuse: Enhancing feature fusion for semantic segmentation.
\newblock In {\em Proceedings of the European Conference on Computer Vision
  (ECCV)}, pages 269--284, 2018.

\bibitem{Zhao2017PyramidSP}
Hengshuang Zhao, J. Shi, Xiaojuan Qi, Xiaogang Wang, and J. Jia.
\newblock Pyramid scene parsing network.
\newblock {\em 2017 IEEE Conference on Computer Vision and Pattern Recognition
  (CVPR)}, pages 6230--6239, 2017.

\bibitem{Zoph2020RethinkingPA}
Barret Zoph, G. Ghiasi, Tsung-Yi Lin, Yin Cui, Hanxiao Liu, E.~D. Cubuk, and
  Quoc~V. Le.
\newblock Rethinking pre-training and self-training.
\newblock {\em ArXiv}, abs/2006.06882, 2020.

\bibitem{nas}
Barret Zoph and Quoc~V Le.
\newblock Neural architecture search with reinforcement learning.
\newblock In {\em ICLR}, 2017.

\bibitem{nasnet}
Barret Zoph, Vijay Vasudevan, Jonathon Shlens, and Quoc~V Le.
\newblock Learning transferable architectures for scalable image recognition.
\newblock In {\em CVPR}, 2018.

\end{thebibliography}
}

\end{document}